\documentclass{article}

\PassOptionsToPackage{numbers, compress}{natbib}

\usepackage[final]{neurips_2023}
\usepackage{amsmath}
\usepackage{amsthm}
\usepackage{stackengine}
\usepackage{graphicx}
\usepackage{caption}
\usepackage{subcaption}
\usepackage{algorithm2e}

\newtheorem{thm}{Theorem}

\usepackage[utf8]{inputenc} 
\usepackage[T1]{fontenc}    
\usepackage{hyperref}       
\usepackage{url}            
\usepackage{booktabs}       
\usepackage{amsfonts}       
\usepackage{nicefrac}       
\usepackage{microtype}      
\usepackage{xcolor}         
\usepackage{algEvalCommands}

\title{The logic of NTQR evaluations of noisy AI agents: 
Complete postulates and logically consistent error correlations}

\author{%
 Andr\'es Corrada-Emmanuel \\
 Data Engines\\
  \texttt{andres.corrada@dataengines.com}
}

\begin{document}

\maketitle

\begin{abstract}
  In his well-known "ship of state" allegory (\textit{Republic}, Book VI, 488) 
  Plato poses a question -- how can a crew of sailors presumed to know little, 
  if anything, about the art of navigation recognize the true pilot among them? This question is shown to have a parallel in the political domain: how can a group of citizens presumed to know little, 
  if anything, about the art of governing recognize the true leader among them? The allegory is intended to show that a simple majority voting procedure cannot safely determine who is most qualified to pilot a ship or a ship of state when the voting members are ignorant or biased.
  Are there any voting algorithms for evaluation that would satisfy Plato's safety concerns under such conditions? The answer to this question has relevance for a
  contemporary problem in AI 
  safety: namely how to validate noisy machine learning algorithms in near real-time so as to operate them safely. We formalize these safety concerns by considering the problem of how to grade tests for which we have no answer key. 

  Any algorithm that evaluates other AI agents in an unsupervised setting would be
  subject to the evaluation dilemma - how would we know the evaluation algorithm
  was correct itself? This endless validation chain exists whenever an evaluation
  algorithm encodes information other than that from the observed test responses.
  Conversely, the validation is finite whenever we consider purely algebraic functions
  of observed responses by the algorithms. Complete sets of postulates can be
  constructed that can prove or disprove the logical consistency of any unsupervised
  evaluation algorithm.

  A complete set of postulates exists whenever we are evaluating $N$ experts
  that took $T$ tests with $Q$ questions with $R$ responses each, one for
  each point \ntqrpt in \ntqrs space. Here we discuss the case of evaluating
  binary classifiers that have taken a single test extensively - 
  the $(N,T=1,Q,R=2)$ tests. We show how some of the postulates have been
  previously identified in the ML literature but not recognized as such -
  the \textbf{agreement equations} of Platanios(\cite{Platanios2014}\cite{Platanios2016}).
  The complete postulates for pair correlated binary classifiers are considered
  and we show how it allows for error correlations to be quickly calculated
  once we are given the individual evaluations of the pair.
  We construct an exact evaluator based on the assumption that the classifiers in the
  ensemble are making independent errors and compare it with grading by majority
  voting on three evaluations using the \uciadult and 
  and \texttt{two-norm} datasets. Throughout, we demonstrate how the formalism
  of logical consistency via algebraic postulates of evaluation
  can help increase the safety of machines using AI algorithms.
\end{abstract}

\section{The evaluation dilemma: infinite safety validation chains}

The deck of the Ship of Fools in Plato's \textit{Republic} illustrates a basic 
engineering problem encountered by ignorant monitors of noisy experts. In the allegory, 
the ship owner (the principal) must delegate control to the best pilot from among the 
crew members, all of whom noisily insist that they are best qualified to do the job. 
However, as Plato sets up the scene, it is doubtful whether any of the crew actually 
knows what this entails. The problem is further compounded by the fact that the owner's 
knowledge of navigation is "not much better" than anyone else on board.
Plato's safety concerns can be restated in an engineering context by considering the 
problem of how to grade a test for which we have no answer key. In our thought 
experiment, the noisy agents are given a test to assess their knowledge of navigation 
The question is whether it is possible to rank or grade the agents using only their 
responses to the test and perhaps ancillary knowledge about the application domain
for the agents. And could we detect, as Plato was worried about, if
they were correlated in their errors.

This problem started to get attention in the ML literature
after the Dawid and Skene 1979 paper\cite{Dawid79} that
considered how to rate the correctness of human doctors
reviewing patient files. Two approaches have been pursued
in the literature since then. The first approach started
with Raykar et al\cite{Raykar2010} and their use of Bayesian
probabilistic methods (). The other used spectral methods and
was started by Parisi et al\cite{Parisi1253}. What is common
to all these methods is their use of probability theory and
the need to train or set the parameters for their use.

This creates an infinite chain of validation that is commonly
known in the Bayesian world as - "it's turtles all the way
down." These probabilistic methods may very well be completely
correct when we use them. But verifying this correctness is
essential for safety to occur when we act on those evaluations.
Bayesian methods, for example, have hyperparameters that model
the expected error of test takers. To be truly safe we would
then need to check whether those hyperparameter settings
applied. This is an infinite validation chain. It can be
represented typographically by,
\begin{quote}
    \ldots( \validatori{2} ( \validatori{1}( grader of experts( experts))))))\ldots
\end{quote}
In this paper we will describe how to build a finite safety validation chain
that can be represented by,
\begin{quote}
    (validator of grader (grader of experts (experts)))
\end{quote}
For any specific \ntqrpt, we can build a complete set of postulates
for the possible evaluations of classifiers given their test responses.
These algebraic postulates relate unknown statistics of the correctness
of the responses by the agents to how they responded in the test. They
are essentially sample statistics of the test. For that reason, since
any given test is finite, we can always build finite representations
that explain completely any data streaming statistic of their decisions
we want to keep.

The focus in this paper will be on the case of binary classifiers or
the $(N,T=1,Q,R=2)$ tests. We will use a dual language treating the
same evaluation with two different viewpoints. One, the ML viewpoint,
is familiar to those that build and test classifiers. But we can
transform our variables and consider the same binary classification
test as a binary response multiple-choice exam. That viewpoint we will
denote as the \ntqrs viewpoint. The variable change results in different
equations for exactly the same postulates in the ML space. The geometry
of evaluation in unsupervised settings is easier to visualize in \ntqrs space.
It should be clear by our variable choices what viewpoint is being used.

The central thesis of the paper is that each \ntqrpt in \ntqrs space
contains a complete set of algebraic postulates relating unknown
statistics of correctness in a test to observables of the aligned
decisions by members in an ensemble of noisy AI algorithms.
The reader will see that some of these postulates are familiar. For
example, in the case of binary response tests each question can be
considered to belong to two types - the `a' type questions and the
`b' type questions. And thus a test of $Q$ questions is going to
satisfy the postulate,
\begin{equation}
    \Qa + \Qb = Q,
\end{equation}
where \Qa denotes the number of `a' questions and \Qb the `b' ones.
By defining the prevalence of the question types (the percentage of
times a question type appeared in the test) by,
\begin{align}
    \prva &= \frac{\Qa}{Q} \\
    \prvb &= \frac{\Qb}{Q},
\end{align}
we can rewrite the postulate in two equivalent forms,
\begin{align}
    \Qa + \Qb &= Q \\
    \prva + \prvb &= 1.
\end{align}

\subsection{Removing semantics from logical consistency of unsupervised monitoring}

To fully validate a set of noisy algorithms
one would want both consistency and soundness. Soundness, however, establishing the
the truth value of the evaluation, cannot be achieved without the answer key. Soundness
requires the truth table of the premises of the evaluation and that is precisely
what the answer key is.

Statistics of correctness in a test need not have any semantic meaning and we
will emphasize that in this paper by talking about the test as a binary
response multiple choice exam. In this viewpoint, all statistics of correctness
are integers. This is the space of the variables \Qa and \Qb.
The other space we can choose to describe exactly the same test is the space
familiar to the ML community when they talk about the problem of binary
classification. A test consists of labeling items in a dataset according to
two labels called \lbla and \lblb in this paper. This is the space of the
variables \prva and \prvb.

In the task of binary classification, we can ascribe a semantic connection
between the label and what items are in the test. If we are classifying
images as dogs or cats, then \prva represents the percentage of dogs in the
images that have been classified. But if we are talking about an arbitrary
test, there need not be any semantic connection between correct response
`a' in one question and correct response `a' in any other question. In that
case, \prva represents the percentage of questions that had correct response
`a' and has no semantic connection to the subject of the test.

\subsection{Digitizing engineering spec and safety concerns}

A binary test as an instrument is separate from the evaluation that
may be under consideration in a general AI machine. But whatever
that evaluation is, it can be compressed into the binary test
format and the evaluation tested for its logical consistency. Take
the case of a nautical test that may involve essay answers. These
could be graded by a ChatGPT-like algorithm that would then
give each respondent's answer a pass/fail mark. These grading sheets
would then be the equivalent of a binary test taken by the respondents.

If a pass/fail grade may seem extreme, one could choose to give
each essay a score between 0 and 100 at integer intervals. This
would be equivalent to the $(N,T=1,Q,R=100)$ tests. There are no
general rules for safety. Whether the $R=2$ format is appropriate
or the $R=100$ is better suited to a specific AI machine cannot
be established absent the application context. In this paper we
will be discussing the $R=2$ to illustrate what should be understood
as concepts easily generalized to the case of $R>2.$

Similar considerations would apply to the concept of engineering spec
in a machine - meeting but not needlessly exceeding the precision
standards for proper operation. Integer values for $R$ digitize the
precision that applies in any given application context.

It is important that the reader keep in mind two different subjects,
the mathematics of the algebraic postulates and their practical
application. They are distinct. As mentioned above, the evaluation
may be a written exam and we may want to check the logical consistency
of the test by digitizing it into the \ntqrs format.

If we digitize the responses of thermometers monitoring a car engine
into degrees Centigrade from, say, $-50$ to $150$. We would be
losing the physical relation between $R=49$ versus $R=102$. Nonetheless,
there would remain a component of the thermometer readings in the
chosen \ntqrpt format that can be validated logically.

\subsection{Paper outline}

The paper will use the case of evaluating binary classifiers to
discuss the complete evaluation postulates that can be build
for different \ntqrpt tests. We start with the case of the trivial
ensemble ($N=1$) to show how these postulates are already probably
known to the technically disposed readers. We then discuss the
case of error correlated pairs of binary classifiers. We use
the complete postulates discussed here to show a mistake in the
ML literature for the evaluation of error independent classifying
functions by Platanios et al\cite{Platanios2014, Platanios2016}.
Finally we discuss three evaluations with trios of binary classifiers
to show how we can detect highly correlated ensembles of binary
classifiers.

\section{Grading an individual binary classifier}

Stripping all semantics from \ntqrs tests allows one to devise
universal equations that relate statistics of correctness in
the test to observed responses to the questions. In a binary
test we would want to know how many \lbla and \lblb questions got
answered correctly. We denote these integer quantities
as, \Raa{i} and \Rbb{i}, respectively for classifier $i.$ And
it follows from their definition that these numbers cannot
exceed the question types,
\begin{align}
    0 &\leq \Raa{i} \leq \Qa \\
    0 &\leq \Rbb{i} \leq \Qb.
\end{align}
In other words, there cannot be more correct responses for a
question type than its total number.

\begin{figure}[ht]
  \centering
  \includegraphics[width= 3 in]{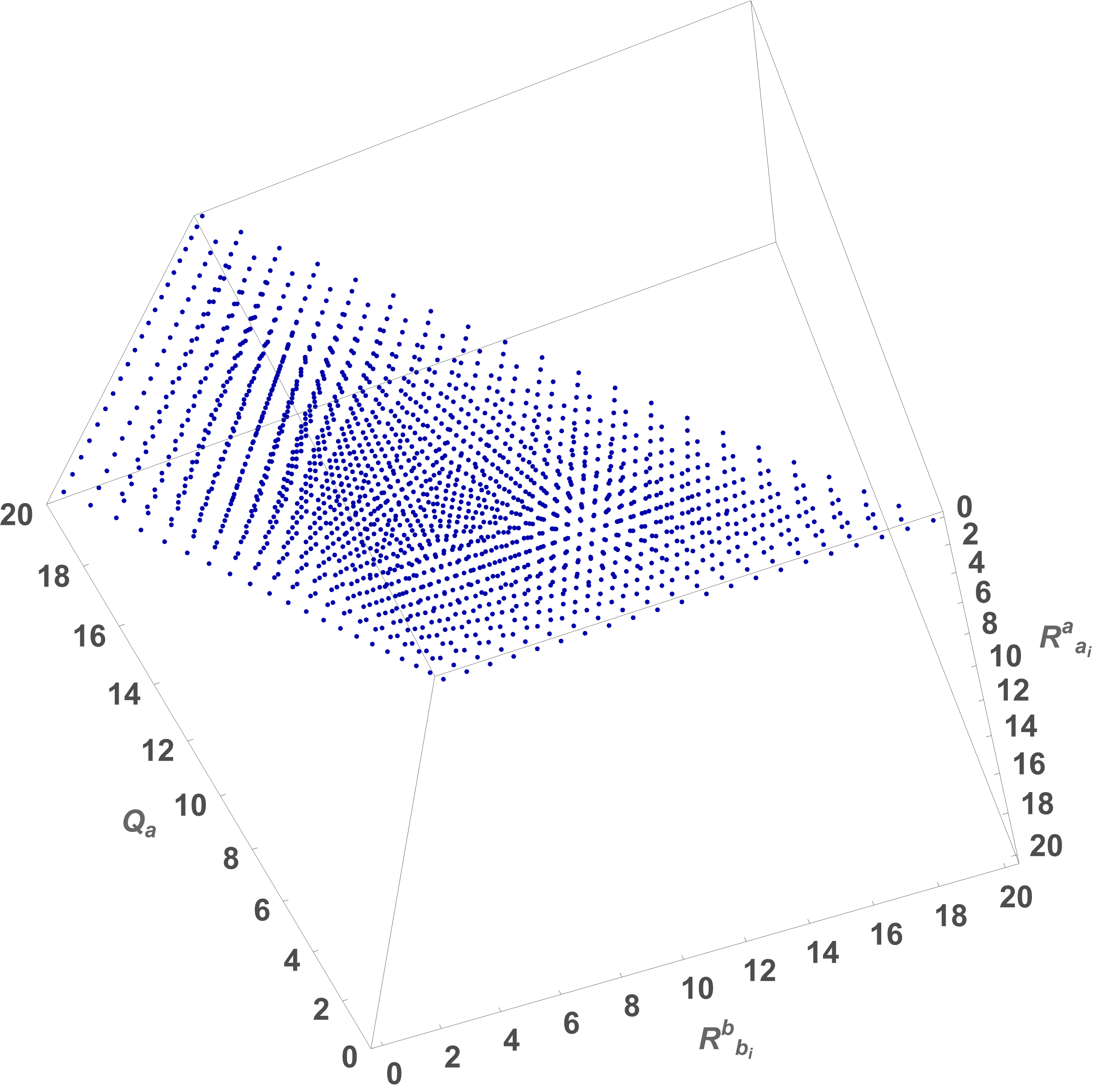}
  \caption{The set of all possible "grades" for a single binary classifier
  that labeled 20 items (the $Q=20$ test). \Qa denotes the number of questions
  whose correct response was "a". The quantities \Rab{i} and \Rbb{i} count
  how many correct responses there were for the \Qa and \Qb ($\Qb=Q-\Qa$)
  questions. The wedge geometry of the possible values in $(\Qa, \Raa{i}, \Rbb{i})$
  space is due to the postulates that connect these variables to each other for
  any test given to a binary classifier.}
  \label{fig:test-apriori-cube}
\end{figure}

We can use these numerical relations to construct the set of all
possible evaluations for a binary test with $Q$. An \emph{evaluation}
of a single binary classifier will be denoted by the tuple,
\begin{equation}
    \binEvali{i},
\end{equation}
of the number of \lbla questions, \Qa, and the number of correct responses
to \lbla and \lblb questions, \Rai{i} and \Rbi{i}, respectively.

Irrespective of whether the labels \lbla and \lblb have semantic meaning,
the number of correct responses in the binary test is given by,
\begin{equation}
    g_i = \Raa{i} + \Rbb{i}.
\end{equation}
We will see eventually that all the difficulties and ambiguities of unsupervised
monitoring arise from the fact that observing test responses gives us a reading
of the \emph{difference} in correct responses,
\begin{equation}
    \Raa{i} - \Rbb{i},
\end{equation}
not their sum as we would wish to know to properly rank them by performance on
the test.

We can use all these relations between $Q$, \Qa, \Qb, \Rai{i}, and \Rbi{i}
to construct the set of all possible evaluations for a binary test of size
$Q.$ That is given by the following algorithm, 
\begin{algorithm}[H]
\SetAlgoLined
\KwResult{All possible evaluations in a binary response test with $Q$ questions.}
evaluations $\leftarrow [\,]$ \tcp{Initialize possible evaluations with the empty list.}
\For{$\Qa \leftarrow 0$ \KwTo $Q$}{
  \For{$\Raa{i} \leftarrow 0$ \KwTo \Qa}{
    \For{$\Rbb{i} \leftarrow 0$ \KwTo $Q - \Qa$}{
      evaluations.append($(\Qa, \Raa{i}, \Rbb{i})$)
    }
  }
}
\end{algorithm}
Figure \ref{fig:test-apriori-cube}
shows the evaluations possible for a $Q=20$ test. In general, before we observe
tests responses, there are $1/6(Q+1)(Q+2)(Q+3)$ possible evaluations. The
experiments we will be considering here will consist of $Q=1000$ tests with
hundreds of millions of possible evaluations. But observing the test
responses will greatly reduce the number of logically consistent
evaluations as we now demonstrate.

\section{Reducing possible evaluations after observing test responses}

We now detail the complete postulates for generating the test responses
of a single binary classifier. This is complete set corresponding to
the $(N=1,T=1,Q,R=2)$ tests. For a single binary classifier we can
observe the number of times a label is assigned. In \ntqrpt space this is
denoted by the number of times we record \lbla and \lblb responses,
\Rai{i} and \Rbi{i} respectively. By logic of the excluded middle for
binary selection, we must have the following two equations as also true
for any evaluation of a binary test,
\begin{align}
    \Rai{i} &= \Raa{i} + \Rab{i} \\
    \Rbi{i} &= \Rba{i} + \Rbb{i}
\end{align}
These postulates are complete because they satisfy,
\begin{equation}
    \Rai{i} + \Rbi{i} = Q.
\end{equation}
There are no more equations that are needed to explain all possible observations of
how a single binary classifier responded during the test.

The same postulates can be rewritten in ML space as,
\begin{align}
    \fai{i} &= \prva \prsa{i} + \prvb \left(1 - \prsb{i} \right) \\
    \fbi{i} &= \prva \left( 1 - \prsa{i} \right) + \prvb \prsb{i}.
\end{align}
These postulates are complete but not independent. They can be expressed
as the single postulate,
\begin{align}
\begin{split}
    \label{eq:single-binary-postulate}
    \prva \left( \prsa{i} -\fai{i} \right) &= \prvb \left( \prsb{i} -\fbi{i} \right) \\
    \Raa{i} - \Rbb{i} &= \Rai{i} - \Rbi{i}
\end{split}
\end{align}

Postulate \ref{eq:single-binary-postulate} causes an enormous reduction in the
number of possible evaluations for binary classifiers.

Consider a series of tests that we carried out on three different datasets well known
in the ML community - \uciadult, \mushroom and \twonorm. These datasets are used
to benchmark the performance of different binary classification algorithms. Here
we will be using them to carry out single tests with 1,000 items classified -
the $(N,T=1,Q=1000,R=2)$ tests. A $Q=1000$ test can have,
\begin{equation}
    \frac{1}{6}(1000 + 1) ( 1000 + 2 ) (1000 + 3 ) = 167,668,501
\end{equation}
possible evaluations. But use of postulate \ref{eq:single-binary-postulate}
reduces the logically consistent evaluations to about a quarter of a million.
The exact number varies on how the classifier responded to the test. We show
the number of evaluations per possible \Qa value in the Figures below. Alongside
the count of possible evaluations, we also plot the plane described by
postulate \ref{eq:single-binary-postulate}. Note that in \ntqrs variables,
postulate \ref{eq:single-binary-postulate} is linear (a plane). But in
ML evaluation space, it is a quadratic surface.

\begin{figure}[ht]
  \centering
  \includegraphics[width=\textwidth]{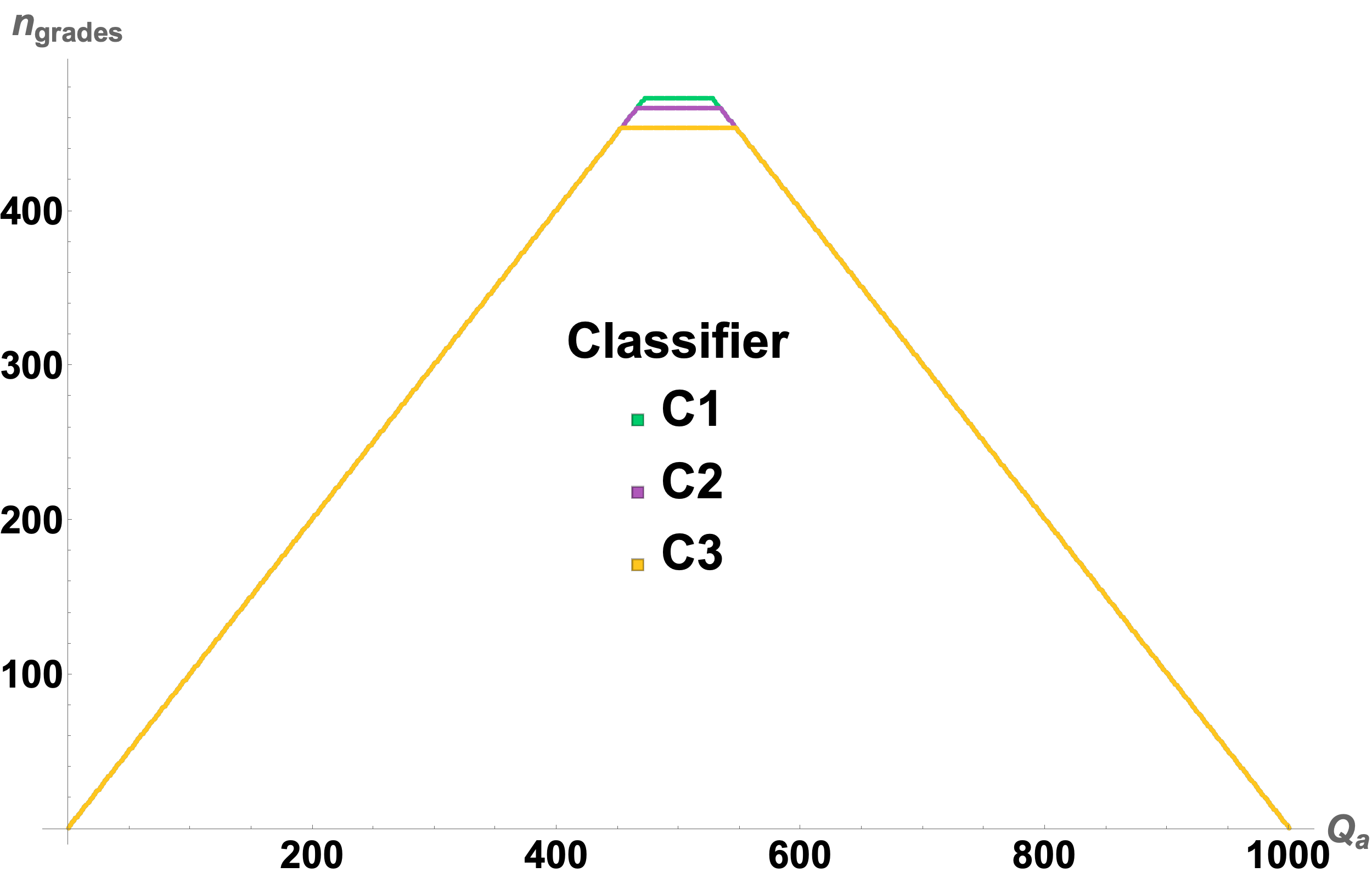}
  \caption{The count of evaluations possible after we observe the labeling
         decisions of three binary classifiers on a \twonorm test of $Q=1,000$ items 
         from the \twonorm dataset.}
         \label{fig:two-norm-eval-posterior-counts}
\end{figure}

\begin{figure}[ht]
  \centering
  \includegraphics[width=\textwidth]{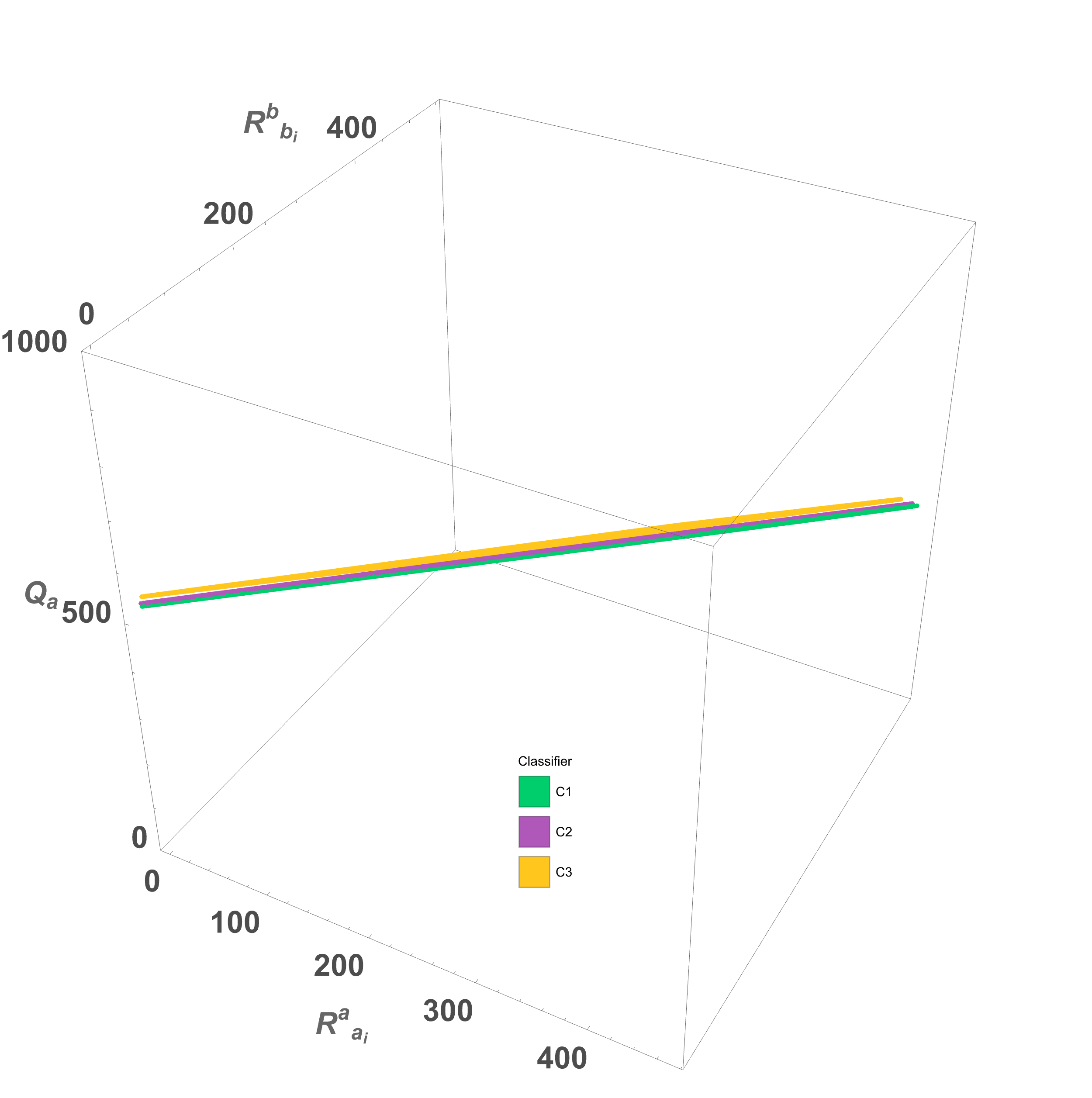}
         \caption{The plane of possible evaluations after we observe the
         aligned decisions of three binary classifiers on a test of 1,000
         items (40/60) drawn from the \twonorm dataset.}
         \label{fig:two-norm-eval-posterior-plane}
\end{figure}

\begin{figure}[ht]
  \centering
  \includegraphics[width=\textwidth]{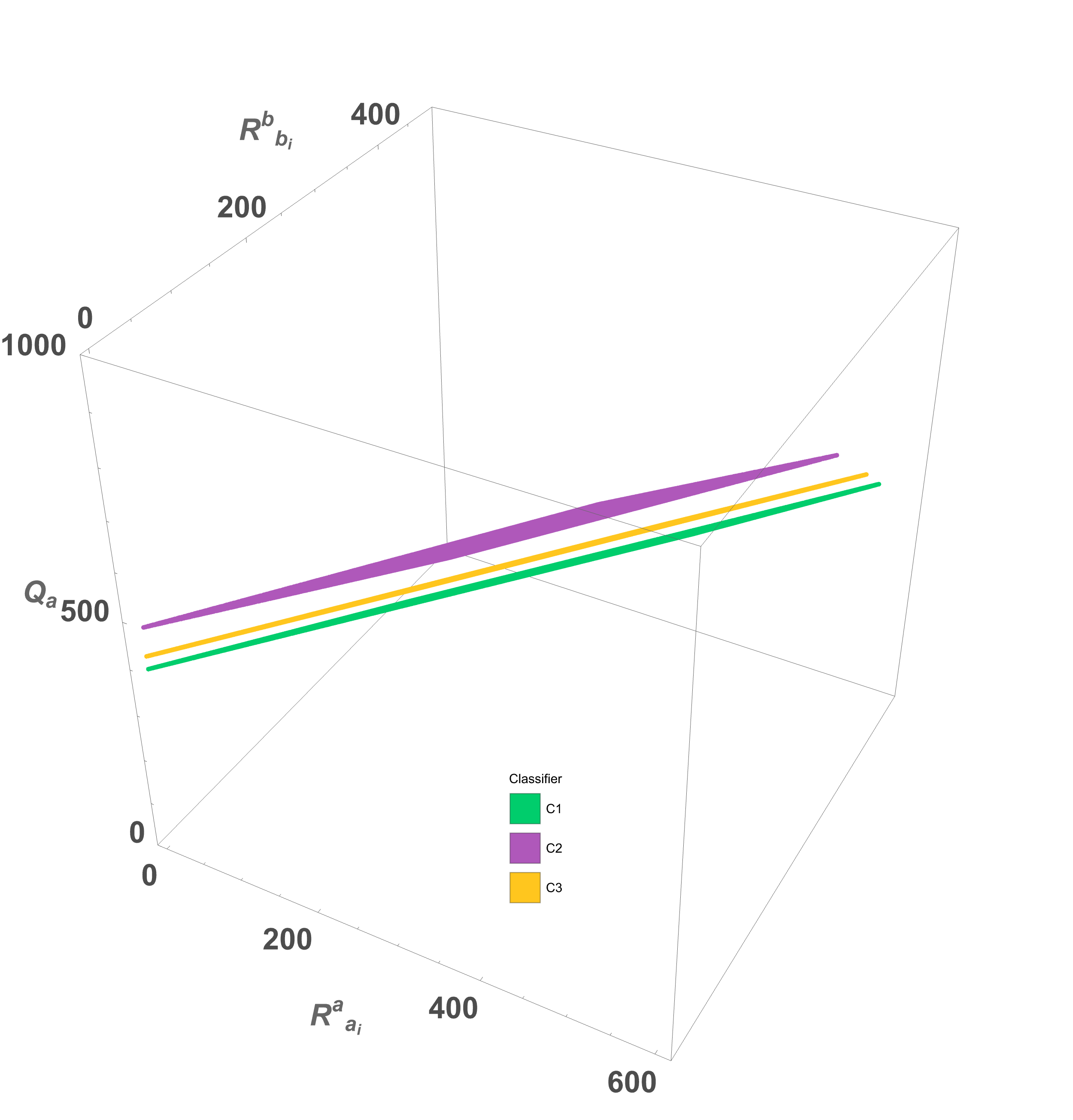}
         \caption{The plane of possible evaluations after we observe the
         aligned decisions of three binary classifiers on a test of 1,000
         items (40/60) drawn from the \uciadult dataset.}
         \label{fig:adult-eval-posterior-plane}
\end{figure}

\section{A pair of binary classifiers and error correlations}

In the previous section we considered the postulates for the $(N=1,T=1,Q,R=2)$
tests where we observe the responses of a single binary classifier. We
now will consider the postulates for the $(N=2,T=1,Q,R=2)$ tests. The
completeness condition in this case will be,
\begin{equation}
    \faaij{i}{j} + \fabij{i}{j} + \fbaij{i}{j} + \fbbij{i}{j} = 1.
\end{equation}
As each item in a dataset is labeled, we keep track of four separate
voting events - the $2^2=4$ ways that two binary classifiers could
be seen to decide on the item's true label.\footnote{It should be
clear that there are many more statistics one may want to consider,
such as consecutive labeling accuracy, etc. This will not be discussed
here.}
What algebraic postulate can we write for a quantity such as \faaij{i}{j}?
Before we do so, we should note that the quantities \faaij{i}{j} and
\fbbij{i}{j} have been discussed together by Platanios et al\cite{Platanios2014,Platanios2016}
as the pair agreement equation, $a_{\{i,j\}}$, in their notation,
\begin{equation}
    a_{\{i,j\}} = \faaij{i}{j} + \fbbij{i}{j}.
\end{equation}
They did not call the agreement equations postulates, but that is
what they are. They noted that their derivation had been purely
algebraic and used rules of logic alone. They also noted that similar
equations underlined the work of previously cited references when they
considered a probabilistic treatment to unsupervised monitoring.

The algebraic treatment of Platanios et al\cite{Platanios2014,Platanios2016},
however, is flawed. It treats the agreement equations as if they were
a system of linear equations that can be understood by variable and equation
counting. This is incorrect. The postulates are not independent of each other.
And given their multi-dimensional nature - they are not just linear equations -
equation and variable counting are not available.

Unfortunately, nobody has noticed that Platanios et al offered a solution
to the problem of error independent classifiers that is incorrect. It is
worthwhile to go into their derivation because there is much about
their approach that highlights the utility of using algebraic postulates
in unsupervised monitoring.

They note that after some algebraic manipulation, they can write the
agreement rate between two classifiers in terms of their individual
error rates ($e_{i}$ and $e_{j}$) as well as their pair error rate,
$e_{i,j}$ as so,
\begin{equation}
    a_{i,j} = 2 - e_i - e_j + 2 e_{i,j}
\end{equation}
They then note, incorrectly, that for error independent classifiers,
\begin{equation}
    e_{i,j} = e_i e_j.
\end{equation}
Their error was not visible to the reviewers and it is not visible to
the reader now. The assumption of product separation seems innocent enough.
But it is only when one attempts to write down exactly what \pei{i}, \pei{j}
and \peij{i}{j} mean in term of the complete postulates that one can see
the error. We will do that in due time. For now we want to proceed with
a description of the Platanios derivation. A solution to the problem of
independent classifiers is then proposed whenever we have three or more
of them. They note that three classifiers provide three readings for
the agreement rates and we only have three quantities, \pei{i}, \pei{j}, and \pei{k} that need to be estimated. This leads to their independent solution
\begin{equation}
    \label{eq:platanios-first}
    \pei{i} = \frac{c \pm \left( 1 - 2 \aij{j}{k} \right)}{\pm 2 \left( 1 - 2 \aij{j}{k} \right)}.
\end{equation}
The quantity $c$ is one that signals that this equation could be wrong,
\begin{equation}
    \label{eq:c-equation}
    c = \sqrt{(1-\aij{i}{j}) (1-\aij{i}{k}) (1-\aij{j}{k})}.
\end{equation}
The quantities \pei{i} are integer ratios for any finite test. If the
three classifiers are error independent, we would have three different
integer ratio products form a perfect square so that equation \ref{eq:c-equation}
would not produce an irrational answer. This is most certainly wrong.
But note how we have detected that something may be wrong - by the appearance
of an unresolved square root. Irrational numbers have alerted us that this
equation may be wrong. We will return to this important point later.

\subsection{The complete postulates for correlated binary classifiers}

To deal with pairs of classifiers we have to introduce a new statistic of
the correctness of the classifiers that captures their error correlations.
This quantity is defined for both labels as,
\begin{align}
    \corratwo{i}{j} &= \frac{\Raaaij{i}{j}}{\Qa} - \frac{\Raa{i}}{\Qa} \frac{\Raa{j}}{\Qa} \\
    \corrbtwo{i}{j} &= \frac{\Rbbbij{i}{j}}{\Qb} - \frac{\Rbb{i}}{\Qb} \frac{\Rbb{j}}{\Qb}
\end{align}

\begin{subequations} \label{eq:pair-correlated-postulates}
\begin{align}
    \faaij{i}{j}& = \prva(\prsa{i} \prsa{j} + \corratwo{i}{j}) + \prvb((1-\prsb{i})(1- \prsb{j}) + \corrbtwo{i}{j}) \\
    \fabij{i}{j}& = \prva(\prsa{i} (1 - \prsa{j}) - \corratwo{i}{j}) + \prvb((1-\prsb{i})\prsb{j} - \corrbtwo{i}{j}) \\
    \fbaij{i}{j}& = \prva((1 - \prsa{i}) \prsa{j} - \corratwo{i}{j}) + \prvb(\prsb{i}(1- \prsb{j}) - \corrbtwo{i}{j}) \\
    \fbbij{i}{j}& = \prva((1 - \prsa{i}) (1 - \prsa{j}) + \corratwo{i}{j}) + \prvb(\prsb{i}\prsb{j} + \corrbtwo{i}{j}).
\end{align}
\end{subequations}

\subsection{The algebraic mistake in the Platanios et al independent solution}

As we have noted, the Platanios independent solution is incorrect. The mistake
will be shown here now that we have the complete postulates for a pair of
binary classifiers. Let us return to the \aij{i}{j} equation,
\begin{equation}
    \aij{i}{j} = 1 - \pei{i} - \pei{j} + 2 \peij{i}{j}.
\end{equation}
The error rate of a single binary classifier, \pei{i} is
just one minus its correctness rate,
\begin{align}
    \pei{i} &= 1 - \prva \prsa{i}  - \prvb \prsb{j} \\
            &= \prva \left( 1 - \prsa{i} \right) + \prvb \left( 1  - \prsb{i} \right)
\end{align}
Using this equation for \pei{i} and the postulate expressions for
\faaij{i}{j} and \fbbij{i}{j} we find that,
\begin{equation}
    \label{eq:correct-pair-error-rate}
    \peij{i}{j} = \left( \prva \left( 1 - \prsa{i} \right) \left( 1 - \prsa{j} \right) +
    \prvb \left( 1  - \prsb{i} \right) \left( 1  - \prsb{j} \right) \right).
\end{equation}
It is clear now what the error by Platanios et al was. It is not the case that we
cannot algebraically factor
\begin{equation}
    \peij{i}{j} = \pei{i}\pei{j},
\end{equation}
when the classifiers are independent. That is, equation \ref{eq:correct-pair-error-rate}
does not generally factor as,
\begin{equation}
\begin{split}
    \left( \prva \left( 1 - \prsa{i} \right) \left( 1 - \prsa{j} \right) +
    \prvb \left( 1  - \prsb{i} \right) \left( 1  - \prsb{j} \right) \right) \neq \\
    \left( \prva \left( 1 - \prsa{i} \right) + \prvb \left( 1  - \prsb{i} \right) \right)
    \left( \prva \left( 1 - \prsa{j} \right) + \prvb \left( 1  - \prsb{j} \right) \right)
    \end{split}
\end{equation}

The Platanios independent solution is incorrect. Nonetheless, it is possible to obtain
a correct independent solution that has many of the good features we want in an
evaluator being used in unsupervised settings. To get the independent binary classifier
solution, we need to consider the complete postulates for the $N=3,T=1,Q,R=2$ tests.
Before we present the error independent algebraic evaluation for those tests,
let us finish our discussion of pair correlated classifiers.

\subsection{An alternative basis for the pair correlated complete postulates}

The postulates for pair correlated binary classifiers (equations \eqref{eq:pair-correlated-postulates})
can be rewritten in a more practical fashion by using tools from algebraic
geometry. As we can see in the equations, the individual performance of the two
classifiers appears at the same time as the new correlation variables that are
needed to have a complete description of all possible aligned decision frequencies.
We can disentangle them by calculating an alternative formulation of the postulates
called the  basis.\cite{Cox}. Its utility in this case is that it
disentangles the postulates involving individual performance from the new correlation
variables. The basis we will use in the experiments in this paper are as follows,

\begin{subequations} \label{eq:gb-pair-postulates}
    \begin{gather}
        \prva \left( \prsa{i} - \fai{i} \right) - \prvb \left( \prsb{i} - \fbi{i} \right) \\
        \prva \left( \prsa{j} - \fai{j} \right) - \prvb \left( \prsb{j} - \fbi{j} \right) \\
        \left( \prsa{i} - \fai{i} \right) \left( \prsb{j} - \fbi{j} \right) - 
        \left( \prsb{i} - \fbi{i} \right) \left( \prsa{j} - \fai{j} \right) \\
        \left( \prsa{i} - \fai{i} \right) \left( \prsb{j} - \fbi{j} \right) + 
        \prva \left( \corratwo{i}{j} - \deltaij{i}{j} \right) +
        \prvb \left( \corrbtwo{i}{j} - \deltaij{i}{j} \right) \\
        \begin{split}
        \left( \prsa{i} - \fai{i} \right) \left( \prsb{j} - \fbi{j} \right)
        \left( \left( \prsa{i} - \fai{i} \right) + \left( \prsb{i} - \fbi{i} \right) \right) + \\
        \left( \prsa{i} - \fai{i} \right)  \left( \corratwo{i}{j} - \deltaij{i}{j} \right) +
        \left( \prsb{i} - \fbi{i} \right)  \left( \corrbtwo{i}{j} - \deltaij{i}{j} \right)
        \end{split} \\
        \begin{split}
        \left( \prsa{i} - \fai{i} \right) \left( \prsb{j} - \fbi{j} \right)
        \left( \left( \prsa{j} - \fai{j} \right) + \left( \prsb{j} - \fbi{j} \right) \right) + \\
        \left( \prsa{j} - \fai{j} \right)  \left( \corratwo{i}{j} - \deltaij{i}{j} \right) +
        \left( \prsb{j} - \fbi{j} \right)  \left( \corrbtwo{i}{j} - \deltaij{i}{j} \right).
        \end{split} 
    \end{gather}
\end{subequations}
These six postulates are not all algebraically independent\footnote{This is notion of algebraic
independence is altogether different from the notion of label error independence. There are
many notions of independence in mathematics.} There are only three that are. Two from the first
three equations, and one from the last three.

The quantity \deltaij{i}{j} is defined as,
\begin{align}
    \deltaij{i}{j} &= \fbbij{i}{j} - \fbi{i}\fbi{j} \\
                   &= \faaij{i}{j} - \fai{i}\fai{j}.
\end{align}
Either definition is equivalent. This equivalence signals that pair correlation is the
only interesting case for binary classifiers in terms of resolving what evaluations
are logically consistent with the aligned ensemble decisions. When we consider the
case of three labels then we will have,
\begin{equation}
    \faaij{i}{j} - \fai{i}\fai{j} \neq \fbbij{i}{j} - \fbi{i}\fbi{j} \neq 
    f_{\gamma_i,\gamma_j} - f_{\gamma_i}  f_{\gamma_j}.
\end{equation}
For the case of $(N,T,Q,R=2)$ tests, a single pair observable, \deltaij{i}{j},
is sufficient.

The consideration of the complete postulates for $(N,T,Q,R=2)$ tests has
the practical utility that we now have a single postulate that relates
individual evaluations of a pair of binary classifiers to their possibly
correlated joint responses. The  experimental
results in the next section are going to use the postulate in the form,
\begin{equation}
   \label{eq:pair-correlation-postulate}
    \left( \prsa{i} - \fai{i} \right) \left( \prsb{j} - \fbi{j} \right) + 
        \prva \left( \corratwo{i}{j} - \deltaij{i}{j} \right) +
        \prvb \left( \corrbtwo{i}{j} - \deltaij{i}{j} \right)
\end{equation}
This will allow us to calculate all the error correlations that are 
logically consistent with the observations \textbf{given} 
the individual evaluations of a pair binary classifiers.
As we will see, there is only one error correlation that will be logically
consistent once one decides how a classifier is performing. The utility
of being able to quickly compute the error correlation of binary classifiers
given their performance and responses will be shown in the next section
when we consider the $(N=3,T=1,Q,R=2)$ tests.

\section{The exact algebraic solution for three error independent binary classifiers}

It is impossible to solve the evaluation of error independent binary classifiers with
just statistics of their pair alignments as in the pair agreement equations considered
by Platanios. The error independent binary classifiers ensemble is solved when we keep
track of all possible eight voting patterns for three classifiers,
\begin{equation}
    \completebinarythree
\end{equation}
We have included the eight postulates for the $(N=3,T=1,Q,R=2)$ tests as an appendix.
They are quartic polynomials in seven variables that allow us to rank the classifiers.
But they also contain the variables that encode the error correlations between the classifiers.
For three binary classifiers there are eight of these error correlation variables. One for every
pair and label, and one for every trio and label.

The purpose of the technical debt incurred by ensembles is that we operate them near independence.
Highly correlated ensembles are effectively one defective algorithm. If all ensembles behaved
error independently in their decisions, safety with AI machines would be a much easier task.
Ensemble algorithms like majority voting are perfectly fine in many situations. But when classifiers
start malfunctioning, it does not. Safety requires evaluation methods that can signal their
own errors - detection that their assumptions are wrong. We have made much of how the Platanios
independent solution is wrong, but the algebraic approach used in the agreement equations,
that it is solely based on the observed responses of the classifiers is an important tool
that needs to be used more.

In this section we will illustrate this self-warning feature of algebraic evaluators by considering
the independent solution when it produces irrational outputs. We can briefly sketch the derivation
of the algebraic solution for three error independent binary classifiers as follows. The eight quartic
polynomials result in a descending ladder of variables that terminates in a quadratic polynomial for
the unknown percentage of \lbla questions, \prva. We can denote that as,
\begin{equation}
    a(\ldots)\prva^2 + b(\ldots)\prva + c(\ldots),
\end{equation}
where the `a', `b', `c' factors are complicated polynomials of the frequencies of the
voting patterns by the trio. There are no free parameters in these coefficients that
we can tune. Thus, there is no guarantee that this equation will even have any solutions.
Geometrically we will show various ways that this polynomial will and will not give us
good evaluations for a trio of binary classifiers.

\begin{equation}
   \label{eq;ae-prev}
    \aeprev.
\end{equation}

\subsection{A synthetic evaluation to test two independent evaluators}

\begin{figure}[ht]
  \centering
  \includegraphics[width=\textwidth]{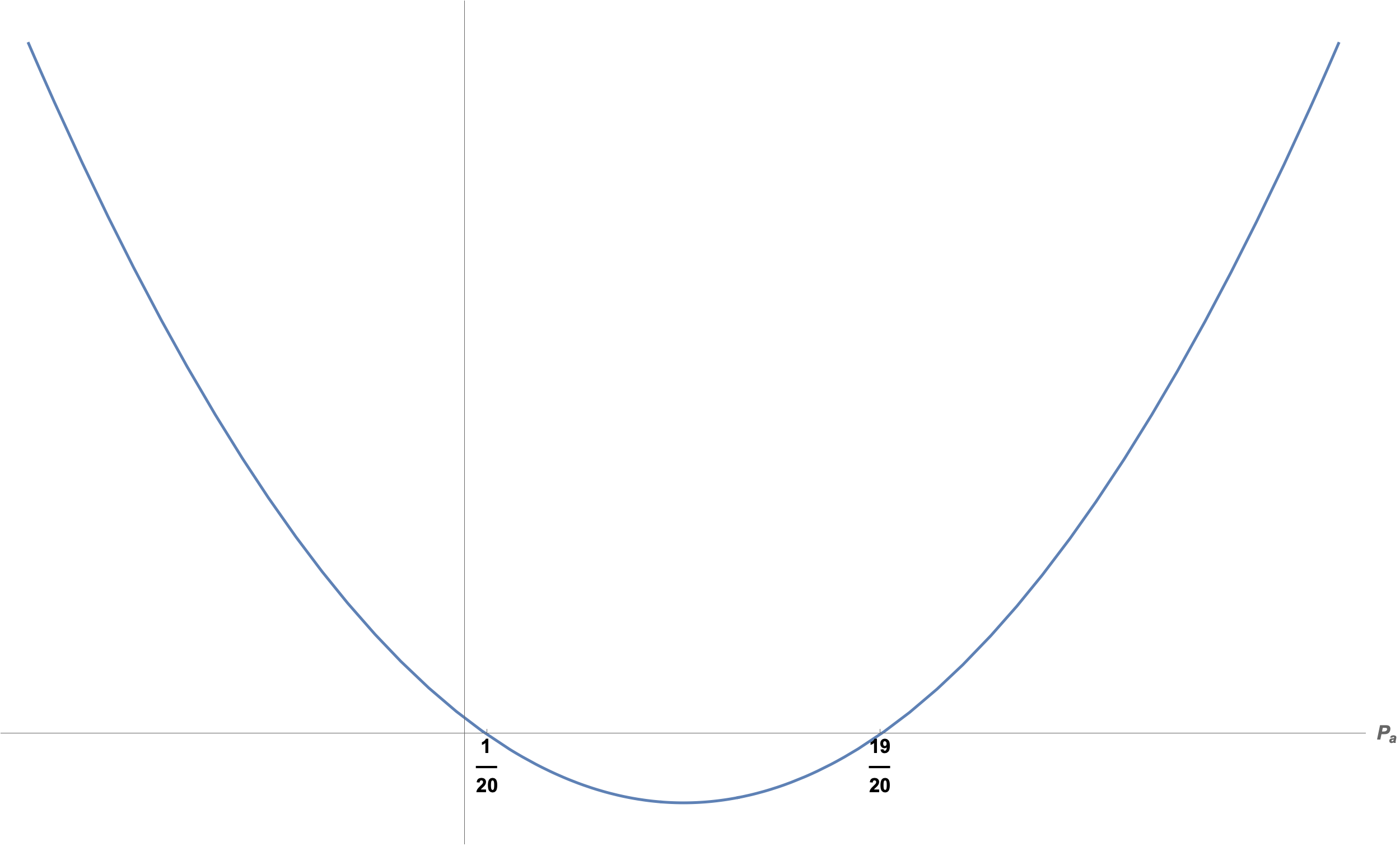}
  \caption{Plot of the \prva quadratic obtained by the error independent
  solution for a trio of binary classifiers. The synthetic evaluation was
  constructed with a value of $\prva = 19/20.$ Since this evaluation was
  done with the assumption of error independence, the recovery of \prva
  is exact, up to logical consistency ($\prva$,  and $1 - \prva$ always
  have the same observed voting frequencies.)}
         \label{fig:synth-prevalence-quadratic}
\end{figure}

Equation \ref{eq;ae-prev} for \prva assumes that the classifiers are
error independent on the test. It has many of the advantages we have
been mentioning for purely algebraic evaluators,
\begin{itemize}
    \item It is purely algebraic.
    \item It has no free parameters other than statistics of how the
    classifiers labeled the items.
\end{itemize}
Like the Platanios independent solution (Equation \ref{eq:platanios-first})
it contains a square root term. As we have emphasized, finite tests have
rational evaluations. Equation \ref{eq;ae-prev} can also be tested by the
same sort of algebraic test we discussed before. In the appendix section
\ref{sec:synth-gen} we detail a synthetic evaluation with independent classifiers.
An independent evaluator cannot have any resolved square roots by construction.
Table \ref{tbl:platanios-vs-ae} details how the square roots resolve. Note
that equation \ref{eq;ae-prev} is more complex than the Platanios solution. But
it resolves into a rational number as it should.

Table
\begin{table}[]
\caption{Comparison of the square root terms for complete postulates independent
evaluator and the Platanios independent solution.}
\label{tbl:platanios-vs-ae}
\centering
\begin{tabular}{@{}cc@{}}\toprule
$\sqrt{(1-\aij{i}{j}) (1-\aij{i}{k}) (1-\aij{j}{k})}$  \\
$\sqrt{\left( 1 - \frac{15961}{25000} \right) 
\left( 1 - \frac{28867}{5000} \right) 
\left( 1 -  \frac{1948}{3125} \right)}$ \\
\addlinespace[3pt]
$\frac{\sqrt{\frac{226746939639}{10}}}{625000}$ \\
\addlinespace[3pt]
\midrule
\sqrtprevterm \\
$\sqrt{4 \left(-\frac{323}{200000} \right)
\left( -\frac{18411}{4000000} \right) 
\left( \frac{1083}{200000} \right) + 
\left( \frac{78413}{5000000} - \left( \frac{543}{2000} \frac{19}{100} \frac{637}{2000} +
\frac{543}{2000} \frac{1083}{200000} +
\frac{19}{100} \left( -\frac{18411}{4000000} \right) +
\frac{637}{2000} \left(-\frac{323}{200000} \right)
\right)
\right)^2}$ \\
\addlinespace[3pt]
$\sqrt{\frac{6440333499}{40000000000000000} +
\left(
\frac{78413}{5000000} - \frac{3302219}{200000000}
\right)}$ \\
$\sqrt{\frac{6440333499}{40000000000000000} +
\frac{27456158601}{40000000000000000}}$ \\
$\sqrt{\frac{338964921}{400000000000000}}$ \\
$\frac{18411}{20000000}$
\end{tabular}
\end{table}

But there is no guarantee that the independent evaluator will resolve
into rational numbers at all. In fact, it can be shown that if equation
\ref{eq;ae-prev} contains an unresolved square root - the classifiers
are provably not error independent on the test. In the following three
sections we will discuss three evaluations. One will go well. One
will cause the independent evaluator to produce an imaginary estimate
for \prva. The last one will go well but the majority voting grader
estimates the classifiers as much more correlated than they are.

\subsection{A \twonorm evaluation}

\begin{figure}[ht]
  \centering
  \includegraphics[width=\textwidth]{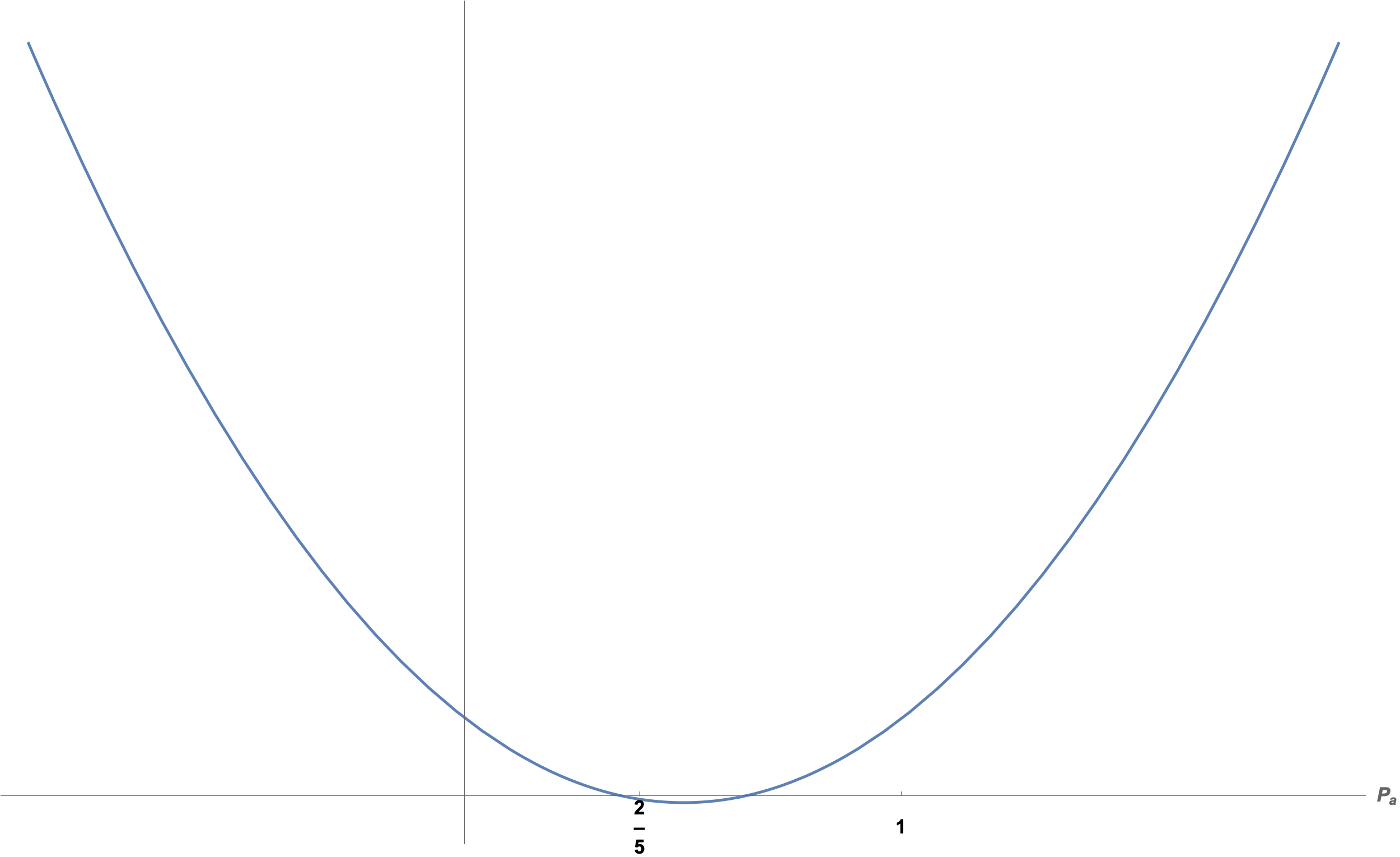}
  \caption{Plot of the \prva quadratic obtained by the error independent
  solution for a trio of binary classifiers. The \twonorm evaluation was
  constructed with a value of $\prva = 4/10.$ \twonorm is also a synthetic
  dataset so the features have very little correlation. Here the algebraic
  independent solution model performs well. It estimates $\prva = 362/1000$,
  a value slightly better than that obtained by majority voting grading, 
  $443/1000.$}
         \label{fig:two-norm-prevalence-quad}
\end{figure}

All three evaluations have exactly the same structure. We tested
1000 items with 400 of \lbla and 600 with \lblb. The \twonorm 
dataset is a synthetic dataset that has a a task distinguishing
between two multi-dimensional Gaussian distributions. Both
the independent evaluator (iAE) and majority voting do well
for this evaluation. The pair correlations have been estimated
using the postulates for the $(N=2,T=1,Q,R=2)$ tests. We
can see that the test error correlations were small and both
methods agree.

\begin{table}[]
\caption{Evaluation results for the \twonorm test, label \lbla.
The test had $\Qa=400$. Algebraic evaluation with the independent
model (iAE) estimated $\Qa=362$, while majority voting (MV) estimated $\Qa=443.$}
  \label{tbl:twonorm-a}
  \centering
\begin{tabular}{@{}lccc@{}}\toprule
C & \multicolumn{3}{c}{\prsa{i}} \\
\cmidrule(lr){2-4}
 & GT & iAE & MV  \\
 \hline
 \addlinespace[3pt]
 1 & $\frac{297}{400}/0.743$ & $\frac{59839+\sqrt{657031321}}{111804}/0.764$ & $\frac{357}{443}/0.806$ \\
 \addlinespace[4pt]
 2 & $\frac{291}{400}/0.728$ & $\frac{56635+\sqrt{657031321}}{106368}/0.773$ & $\frac{357}{443}/0.806$ \\
 \addlinespace[4pt]
 3 & $\frac{293}{400}/0.733$ & $\frac{53317+\sqrt{657031321}}{102096}/0.773$ & $\frac{353}{443}/0.797$ \\
 \addlinespace[4pt]
 \midrule[1pt]
 pair &  \multicolumn{3}{c}{\corratwo{i}{j}} \\
 \cmidrule(lr){2-4}
 & GT & iAE & MV \\
1,2 & $-\frac{427}{160000}/-0.003$ & $-\frac{3485}{65522}/-0.05$ & $-\frac{7396}{196249}/-0.038$ \\
\addlinespace[4pt]
1,3 & $\frac{579}{160000}/0.004$ & $-\frac{3485}{65522}/-0.053$ & $-\frac{7740}{196249}/-0.039$ \\
\addlinespace[4pt]
2,3 & $\frac{1537}{160000}/0.010$ & $-\frac{1681}{32761}/-0.051$ &  $-\frac{7740}{196249}/-0.039$ \\
\addlinespace[4pt]
\end{tabular}
\end{table}

\begin{table}[]
\caption{Evaluation results for the \twonorm test, label \lblb.The 
test had $\Qb=600$. Algebraic evaluation with the independent
model (iAE) estimated $\Qb=638$, while majority voting (MV) estimated $\Qb=557.$}
  \label{tbl:twonorm-b}
  \centering
\begin{tabular}{@{}lccc@{}}\toprule
C & \multicolumn{3}{c}{\prsb{i}}\\
\cmidrule(lr){2-4}
 & GT & iAE & MV \\
 \hline
 \addlinespace[3pt]
 1 & $\frac{17}{24}/0.708$ & $\frac{51965+\sqrt{657031321}}{111804}/0.694$ & $\frac{442}{557}/0.794$ \\
 \addlinespace[4pt]
 2 & $\frac{17}{24}/0.708$ & $\frac{49733+\sqrt{657031321}}{106368}/0.709$ & $\frac{448}{557}/0.804$ \\
 \addlinespace[4pt]
 3 & $\frac{11}{15}/0.733$ & $\frac{48779+\sqrt{657031321}}{102096}/0.729$ & $\frac{457}{557}/0.820$ \\
 \addlinespace[4pt]
 \midrule[1pt]
 pair & \multicolumn{3}{c}{\corrbtwo{i}{j}}\\
 \cmidrule(lr){2-4}
 & GT & iAE & MV  \\
1,2 & $\frac{1309}{21312}/0.061$ & $\frac{6109}{203522}/0.030$ & $-\frac{12535}{310249}/-0.040$ \\
\addlinespace[4pt]
1,3 & $\frac{1}{450}/0.006$ & $\frac{12201}{407044}/0.030$ & $-\frac{11500}{310249}/-0.037$ \\
\addlinespace[4pt]
2,3 & $\frac{1}{180}/0.010$ & $\frac{2961}{101761}/0.029$ & $-\frac{8158}{310249}/-0.026$\\
\addlinespace[4pt]
\end{tabular}
\end{table}

\subsection{The \uciadult imaginary evaluation}

\begin{figure}[ht]
  \centering
  \includegraphics[width=\textwidth]{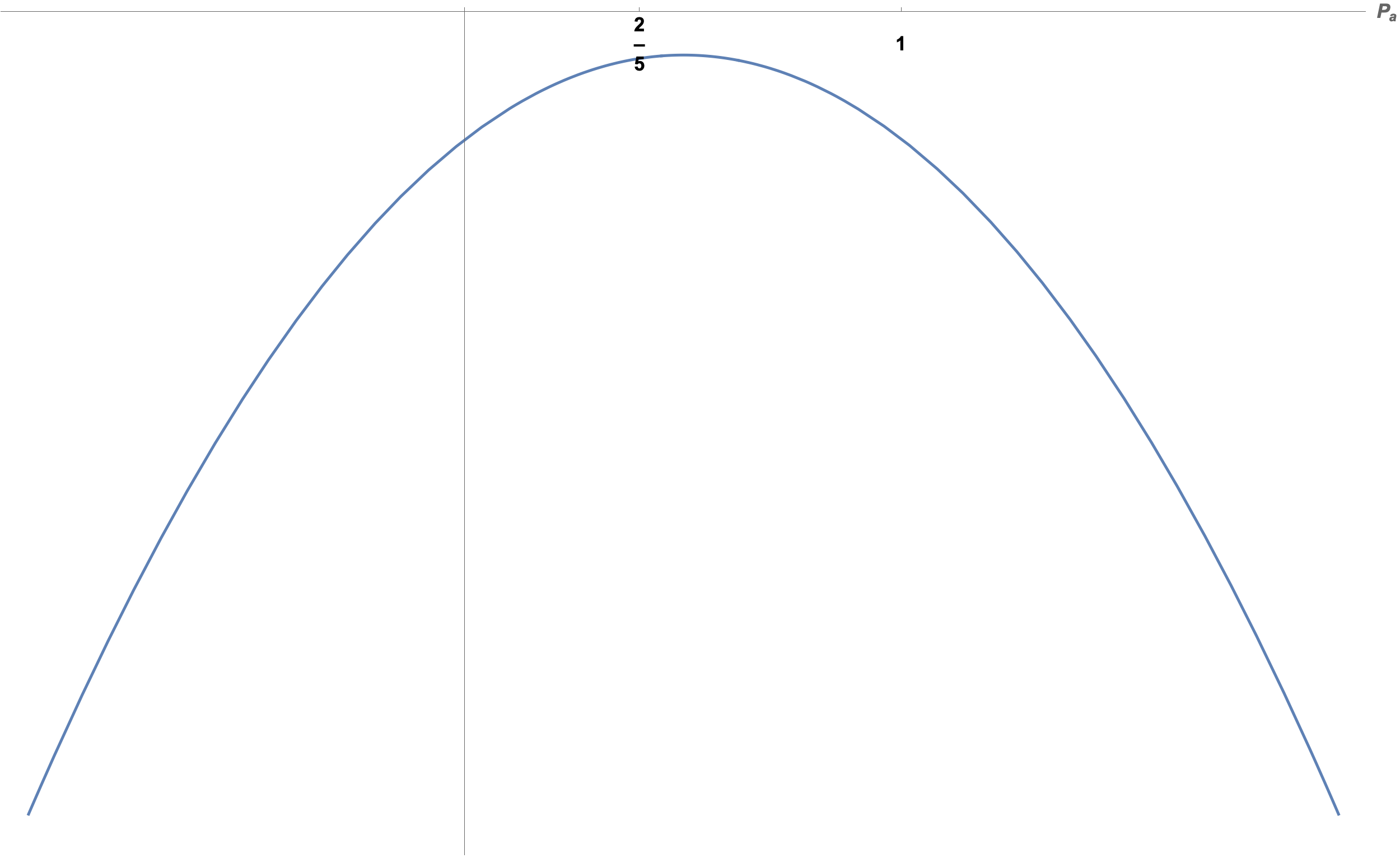}
  \caption{Plot of the \prva quadratic obtained by the error independent
  solution for a trio of binary classifiers. The \uciadult evaluation was
  constructed with a value of $\prva = 4/10.$ \uciadult features are more
  correlated than those in the \twonorm dataset. This starts to degrade
  the estimates from both the iAE and MV algorithms. Here the iAE evaluator
  fails to return a real estimate. As the plot shows, the prevalence
  quadratic has no solution for this test. The MV estimate is $443/1000$, 
  $443/1000.$}
         \label{fig:adult-imag-prevalence-quad}
\end{figure}

We now consider an evaluation on a realistic dataset, \uciadult.
This is a dataset that is extensively used in the fairness in
ML literature as it contains sensitive attributes such as race and gender.
Here the error correlation between classifiers 1 and 2 is large
enough that it triggered the independent evaluator to produce an
imaginary number. We see majority voting also estimates a large
correlation, albeit not for the correct pair.

Tables \ref{tbl:adult-a-imag} and \ref{tbl:adult-b-imag}.

\begin{table}[]
\caption{Evaluation results for the \uciadult test, label \lbla.
The test had $\Qa=400$. Algebraic evaluation with the independent
model (iAE) estimated $\Qa=362$, while majority voting (MV) estimated $\Qa=582.$}
  \label{tbl:adult-a-imag}
  \centering
\begin{tabular}{@{}lccc@{}}\toprule
C & \multicolumn{3}{c}{\prsa{i}} \\
\cmidrule(lr){2-4}
 & GT & iAE & MV  \\
 \hline
 \addlinespace[3pt]
 1 & $\frac{67}{100}/0.67$ & -- & $\frac{357}{582}/0.689$ \\
 \addlinespace[4pt]
 2 & $\frac{157}{200}/0.785$ & -- & $\frac{357}{582}/0.838$ \\
 \addlinespace[4pt]
 3 & $\frac{363}{400}/0.908$ & -- & $\frac{353}{582}/0.955$ \\
 \addlinespace[4pt]
 \midrule[1pt]
 pair &  \multicolumn{3}{c}{\corratwo{i}{j}} \\
 \cmidrule(lr){2-4}
 & GT & iAE & MV \\
1,2 & $-\frac{419}{20000}/-0.021$ & -- & $-\frac{8507}{169362}/-0.050$ \\
\addlinespace[4pt]
1,3 & $-\frac{121}{40000}/-0.003$ & -- & $-\frac{2353}{169362}/-0.013$ \\
\addlinespace[4pt]
2,3 & $\frac{4009}{80000}/0.005$ & -- &  $\frac{11482}{84681}/0.136$ \\
\addlinespace[4pt]
\end{tabular}
\end{table}

\begin{table}[]
\caption{Evaluation results for the \uciadult test, label \lblb.The 
test had $\Qb=600$. Algebraic evaluation with the independent
model (iAE) estimated $\Qb=638$, while majority voting (MV) estimated $\Qb=557.$}
  \label{tbl:adult-b-imag}
  \centering
\begin{tabular}{@{}lccc@{}}\toprule
C & \multicolumn{3}{c}{\prsb{i}}\\
\cmidrule(lr){2-4}
 & GT & iAE & MV \\
 \hline
 \addlinespace[3pt]
 1 & $\frac{263}{600}/0.438$ & -- & $\frac{214}{418}/0.512$ \\
 \addlinespace[4pt]
 2 & $\frac{33}{50}/0.66$ & -- & $\frac{388}{418}/0.928$ \\
 \addlinespace[4pt]
 3 & $\frac{77}{120}/0.642$ & -- & $\frac{396}{418}/0.947$ \\
 \addlinespace[4pt]
 \midrule[1pt]
 pair & \multicolumn{3}{c}{\corrbtwo{i}{j}}\\
 \cmidrule(lr){2-4}
 & GT & iAE & MV  \\
1,2 & $\frac{1309}{21312}/0.061$ & -- & $-\frac{1530}{43681}/-0.035$ \\
\addlinespace[4pt]
1,3 & $\frac{1}{450}/0.006$ & -- & $-\frac{102}{3971}/-0.0256$ \\
\addlinespace[4pt]
2,3 & $\frac{1}{180}/0.010$ & -- & $\frac{141}{361}/0.391$ \\
\addlinespace[4pt]
\end{tabular}
\end{table}

\subsection{A successful \uciadult evaluation}

\begin{figure}[ht]
  \centering
  \includegraphics[width=\textwidth]{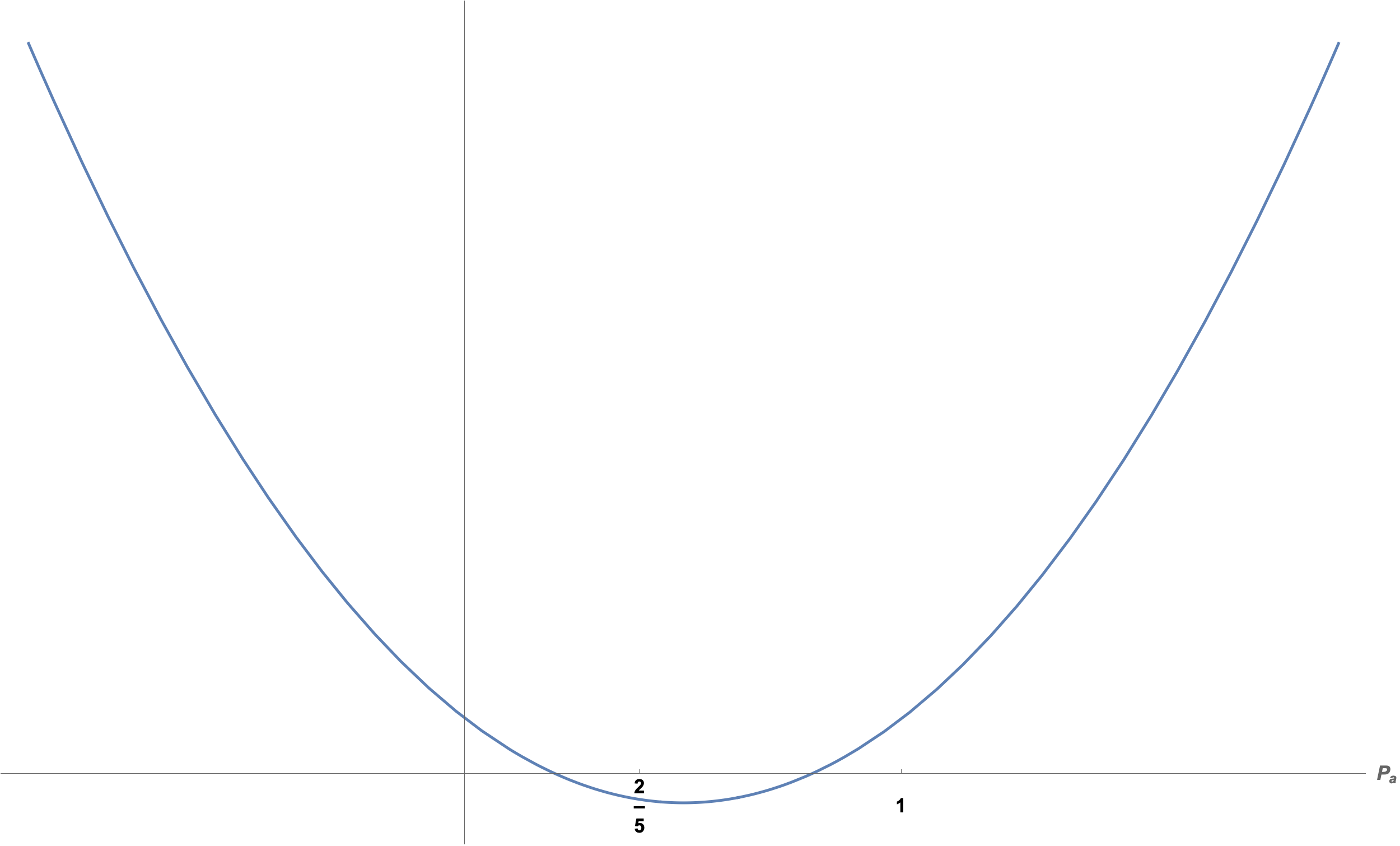}
  \caption{Plot of the \prva quadratic obtained by the error independent
  solution for a trio of binary classifiers. The \uciadult evaluation was
  constructed with a value of $\prva = 4/10.$ This \uciadult evaluation
  gave seemingly correct estimates for iAE that were projected to
  the nearest logically consistent integer values. Here the iAE evaluator
  is starting to give bad prevalence estimates. In this evaluation it
  estimade $\prva = 211/1000$. The MV estimate is better at $447/1000.$
  But the iAE estimates of classifier performance and logically consistent
  error correlations are better than MV. The prevalence quadratic here
  is $\frac{11686017}{250000000000} \prva \left(1  - \prva \right) + \frac{475733909}{61035156250000}.$}
         \label{fig:adult-real-prevalence-quad}
\end{figure}

We can have good evaluations with the \uciadult dataset. Here is one.
Note that the independent AE evaluator is a better estimator of performance
and correlations, but not of prevalence.

Tables \ref{tbl:adult-a-good} and \ref{tbl:adult-b-good}.

\begin{table}[]
\caption{Evaluation results for the \uciadult test, label \lbla.
The test had $\Qa=400$. Algebraic evaluation with the independent
model (iAE) estimated $\Qa=211$, while majority voting (MV) estimated $\Qa=538.$}
  \label{tbl:adult-a-good}
  \centering
\begin{tabular}{@{}lccc@{}}\toprule
C & \multicolumn{3}{c}{\prsa{i}} \\
\cmidrule(lr){2-4}
 & GT & iAE & MV  \\
 \hline
 \addlinespace[3pt]
 1 & $\frac{81}{100}/0.81$ & $\frac{8309+\sqrt{11686017}}{12572}/0.934$ & $\frac{447}{538}/0.831$ \\
 \addlinespace[4pt]
 2 & $\frac{111}{200}/0.555$ & $\frac{5953+\sqrt{11686017}}{11776}/0.796$ & $\frac{314}{538}/0.584$ \\
 \addlinespace[4pt]
 3 & $\frac{7}{8}/0.875$ & $\frac{44407+\sqrt{11686017}}{52648}/0.910$ & $\frac{509}{538}/0.946$ \\
 \addlinespace[4pt]
 \midrule[1pt]
 pair &  \multicolumn{3}{c}{\corratwo{i}{j}} \\
 \cmidrule(lr){2-4}
 & GT & iAE & MV \\
1,2 & $-\frac{291}{20000}/-0.015$ & $-\frac{602}{44521}/-0.014$ & $-\frac{5096}{72361}/-0.070$ \\
\addlinespace[4pt]
1,3 & $-\frac{1}{800}/-0.001$ & $-\frac{266}{44521}/-0.006$ & $-\frac{2639}{289444}/-0.009$ \\
\addlinespace[4pt]
2,3 & $\frac{11}{1600}/0.007$ & $-\frac{817}{44521}/-0.018$ & $-\frac{74193}{289444}/-0.256$ \\
\addlinespace[4pt]
\end{tabular}
\end{table}

\begin{table}[]
\caption{Evaluation results for the \uciadult test, label \lblb.The 
test had $\Qb=600$. Algebraic evaluation with the independent
model (iAE) estimated $\Qb=781$, while majority voting (MV) estimated $\Qb=462.$}
  \label{tbl:adult-b-good}
  \centering
\begin{tabular}{@{}lccc@{}}\toprule
C & \multicolumn{3}{c}{\prsb{i}}\\
\cmidrule(lr){2-4}
 & GT & iAE & MV \\
 \hline
 \addlinespace[3pt]
 1 & $\frac{7}{10}/0.7$ & $\frac{482}{789}/0.611$ & $\frac{214}{462}/0.877$ \\
 \addlinespace[4pt]
 2 & $\frac{121}{150}/0.807$ & $\frac{619}{789}/0.784$ & $\frac{462}{418}/0.948$ \\
 \addlinespace[4pt]
 3 & $\frac{6}{25}/0.24$ & $\frac{175}{789}/0.221$ & $\frac{396}{462}/0.357$ \\
 \addlinespace[4pt]
 \midrule[1pt]
 pair & \multicolumn{3}{c}{\corrbtwo{i}{j}}\\
 \cmidrule(lr){2-4}
 & GT & iAE & MV  \\
1,2 & $\frac{1309}{21312}/0.061$ & $\frac{2251}{622521}/0.004$ & $-\frac{1530}{43681}/-0.035$ \\
\addlinespace[4pt]
1,3 & $\frac{1}{450}/0.006$ & $\frac{862}{622521}/0.001$ & $-\frac{102}{3971}/-0.0256$ \\
\addlinespace[4pt]
2,3 & $\frac{1}{180}/0.010$ & $\frac{2924}{622521}/0.005$ & $\frac{141}{361}/0.391$ \\
\addlinespace[4pt]
\end{tabular}
\end{table}

\section{Conclusions and future work}

Detailed discussions of the $(N=1,T=1,Q,R=2)$, $(N=2,T=1,Q,R=2)$, and $(N=3,T=1,Q,R=2)$
tests have been used to illustrate the idea of complete algebraic evaluation postulates.
We have used them to derive an algebraic evaluator for error independent binary classifiers
that can self-alarm when the classifiers are correlated. In addition, we have used the
algebraic postulates for pairs of binary classifiers to devise a simple algorithm that
selects the only logically consistent pair correlation between binary classifiers once
we are told their individual performance. We used this estimation of error correlation
between classifiers to detect when evaluations should be suspected as having failed.

It is important that users of these algebraic postulates not fall into the metaphysical
mistake of assuming that tells the monitoring anything other than a statistics
of the correctness of the test. In addition, we have shown that we can only show
logical consistency for tests. The logical soundness of the evaluations cannot be done
absent an answer key or external signals that are taken as authoritative by a safety
monitoring system.

It should be clear to the reader that the results presented here can be extended
in many possible directions. We have focused solely on the per-item aligned statistics
of how AI algorithms respond. Different complete postulates would be required for
each different statistic once wished to compute via these algebraic considerations.
We are presently working on the exact solution for the $N=3,T=1,Q,R=3$ tests.

\section{Appendix}

The technical details about all the equations presented in this paper can be found
in a recent ArXiv manuscript.\cite{corradaemmanuel2023streaming}

\subsection{The error independent generating polynomials}

The full derivation of the generating set of polynomials for the frequency of
decisions by error independent binary classifiers is contained in the technical
reference. Here we present it for completeness sake as it is used to generate
a synthetic evaluation in the paper that compares equation \ref{eq;ae-prev} to
the Platanios independent equation \ref{eq:platanios-first}.

\begin{thm}
    The data sketch of three independent classifiers is given the following
    complete set.
\begin{flalign}
  \faaaijk{i}{j}{k}  &=  \prva \prsa{i} \prsa{j} \prsa{k} & + & \; (1 - \prva) (1 - \prsb{i}) (1 - \prsb{j}) (1 - \prsb{k})\\
  \faabijk{i}{j}{k}  &=  \prva  \prsa{i} \prsa{j} (1 - \prsa{k}) & + & \;(1 - \prva) (1 - \prsb{1}{\beta}) (1 - \prsb{j}) \prsb{k}\\
  \fabaijk{i}{j}{k}  &=  \prva  \prsa{i}  (1 - \prsa{j})  \prsa{k} & + & \;(1 - \prva) (1 - \prsb{i}) \prsb{j} (1 - \prsb{k})\\
  \fbaaijk{i}{j}{k}  &=  \prva  (1 - \prsa{i}) \prsa{j} \prsa{k} & + & \;(1 - \prva) \prsb{i} (1 - \prsb{j}) (1 - \prsb{k})\\
  \fbbaijk{i}{j}{k}  &=  \prva  (1 - \prsa{i}) (1 - \prsa{j}) \prsa{k} & + & \;(1 - \prva) \prsb{i} \prsb{j} (1 - \prsb{k})\\
  \fbabijk{i}{j}{k}  &=  \prva  (1 - \prsa{i})  \, \prsa{j} \, (1 - \prsa{k}) & + & \; (1 - \prva) \prsb{i} (1 - \prsb{j}) \prsb{k}\\
  \fabbijk{i}{j}{k}  &= \prva  \prsa{i}  (1 - \prsa{j})  (1 - \prsa{k}) & + & \;(1 - \prva) (1 - \prsb{i}) \prsb{j} \prsb{k}\\
  \fbbbijk{i}{j}{k}  &= \prva  (1 - \prsa{i})  (1 - \prsa{j}) (1 - \prsa{k}) & + & \; (1 - \prva) \prsb{i} \prsb{j} \prsb{k}
\end{flalign}
  This generating set defines a non-empty evaluation variety that contains the true evaluation point.
\end{thm}

\label{sec:synth-gen}
\begin{table}[]
\caption{Synthetic evaluation used to test the independent evaluators where
\prsa equals $19/20.$}
  \label{tbl:synth-3-independent}
  \centering
\begin{tabular}{@{}lcc@{}}\toprule
C & \prsa{i} & \prsb{i} \\
\cmidrule(lr){2-3}
 \addlinespace[3pt]
 1 & $\frac{18}{25}$ & $\frac{11}{100}$  \\
 \addlinespace[4pt]
 2 & $\frac{41}{50}$ & $\frac{19}{50}$\\
 \addlinespace[4pt]
 3 & $\frac{71}{100}$ & $\frac{43}{50}$ \\
 \addlinespace[4pt]
\end{tabular}
\end{table}

\begin{table}[]
\caption{Frequency of voting patterns given the performances in table \ref{tbl:synth-3-independent}.}
  \label{tbl:synth-freqs}
  \centering
\begin{tabular}{@{}lcccccccc@{}}\toprule
pattern & \faaaijk{i}{j}{k} & \faabijk{i}{j}{k} & \fabaijk{i}{j}{k} & \fbaaijk{i}{j}{k} &
\fabbijk{i}{j}{k} & \fbabijk{i}{j}{k} & \fbbaijk{i}{j}{k} & \fbbbijk{i}{j}{k} \\
\addlinespace[4pt]
value & $\frac{2010437}{5000000}$ & $\frac{931913}{5000000}$ & $\frac{448913}{5000000}$ & $\frac{776713}{5000000}$ &
$\frac{251237}{5000000}$ & $\frac{330937}{5000000}$ & $\frac{171437}{5000000}$ & $\frac{78413}{5000000}$ 
\end{tabular}
\end{table}

\begin{table}[]
\caption{Agreement rates needed by the Platanios independent solution equation \ref{eq:platanios-first}.
Note that the c factor is equal to $\frac{\sqrt{\frac{226746939639}{10}}}{625000}$}
  \label{tbl:synth-platanios}
  \centering
\begin{tabular}{@{}lccc@{}}\toprule
pair & \faaij{i}{j} & \fbbij{i}{j} & \aij{i}{j} \\
\cmidrule(lr){2-4}
 \addlinespace[3pt]
 1,2 & $\frac{58847}{100000}$ & $\frac{4997}{100000}$ & $\frac{15961}{25000}$ \\
 \addlinespace[3pt]
 1,3 & $\frac{49187}{100000}$ & $\frac{8187}{100000}$ & $\frac{28687}{50000}$ \\
 \addlinespace[3pt]
 2,3 & $\frac{555743}{100000}$ & $\frac{6593}{100000}$ & $\frac{1948}{3125}$ 
\end{tabular}
\end{table}

\begin{table}[]
\caption{Observed voting rates needed by the independent evaluator.}
  \label{tbl:synth-us}
  \centering
\begin{tabular}{@{}lcc@{}}\toprule
classifier & \fai{i} & \fbi{j} \\
\cmidrule(lr){2-3}
 \addlinespace[3pt]
 1 & $\frac{3193587}{5000000}$ & $\frac{1026563}{5000000}$ \\
 \addlinespace[3pt]
 2 & $\frac{2790287}{5000000}$ & $\frac{858263}{5000000}$ \\
 \addlinespace[3pt]
 3 & $\frac{2958587}{5000000}$ & $\frac{1261563}{5000000}$ \\
 \midrule[1pt]
  pair & \multicolumn{2}{c}{$\Delta_{i,j}$} \\
  \addlinespace[4pt]
  1,2 & \multicolumn{2}{c}{$\frac{368188959931}{25000000000000}$} \\
  \addlinespace[3pt]
  1,3 & \multicolumn{2}{c}{$\frac{751676102031}{25000000000000}$} \\
  \addlinespace[3pt]
  2,3 & \multicolumn{2}{c}{$\frac{565497154931}{25000000000000}$}
\end{tabular}
\end{table}

\subsection{Generating a synthetic evaluation to test independent evaluators}

The equations in the previous section can be used to generate whatever
set of frequencies one wanted and it would be guaranteed to represent
that from classifiers error independent on the test. Table \ref{tbl:synth-3-independent}
shows the results of a single random choice for the correctness statistics of
three binary classifiers. By using the generating polynomials, we can compute,
as in Table \ref{tbl:synth-freqs} the frequency of their aligned decisions.
We then detail in Table \ref{tbl:synth-platanios}
how these frequencies turn into the agreement rates needed by the Platanios
independent solution. Table \ref{tbl:synth-us} shows the rates needed
by the iAE estimate of \prsa ,equation \ref{eq;ae-prev}.

\bibliographystyle{unsrtnat}
\bibliography{ntqrpostulates}

\begin{thebibliography}{7}
\providecommand{\natexlab}[1]{#1}
\providecommand{\url}[1]{\texttt{#1}}
\expandafter\ifx\csname urlstyle\endcsname\relax
  \providecommand{\doi}[1]{doi: #1}\else
  \providecommand{\doi}{doi: \begingroup \urlstyle{rm}\Url}\fi

\bibitem[Platanios et~al.(2014)Platanios, Blum, and Mitchell]{Platanios2014}
Emmanouil~Antonios Platanios, E.~A. Blum, and Tom Mitchell.
\newblock Estimating accuracy from unlabeled data: A bayesian approach.
\newblock In \emph{Proceedings of The 33rd International Conference on Machine Learning}, volume~48 of \emph{Proceedings of Machine Learning Research}, pages 1416--1425, New York, New York, USA, 2014.

\bibitem[Platanios et~al.(2016)Platanios, Dubey, and Mitchell]{Platanios2016}
Emmanouil~Antonios Platanios, Avinava Dubey, and Tom Mitchell.
\newblock Estimating accuracy from unlabeled data: A bayesian approach.
\newblock In Maria~Florina Balcan and Kilian~Q. Weinberger, editors, \emph{Proceedings of The 33rd International Conference on Machine Learning}, volume~48 of \emph{Proceedings of Machine Learning Research}, pages 1416--1425, New York, New York, USA, 2016.

\bibitem[Dawid and Skene(1979)]{Dawid79}
P.~Dawid and A.~M. Skene.
\newblock Maximum likelihood estimation of observer error-rates using the em algorithm.
\newblock \emph{Applied Statistics}, pages 20--28, 1979.

\bibitem[Raykar et~al.(2010)Raykar, Yu, Zhao, Valadez, Florin, Bogoni, and Moy]{Raykar2010}
Vikas~C. Raykar, Shipeng Yu, Linda~H. Zhao, Gerardo~Hermosillo Valadez, Charles Florin, Luca Bogoni, and Linda Moy.
\newblock Learning from crowds.
\newblock \emph{Journal of Machine Learning Research}, 11\penalty0 (43):\penalty0 1297--1322, 2010.

\bibitem[Parisi et~al.(2014)Parisi, Strino, Nadler, and Kluger]{Parisi1253}
Fabio Parisi, Francesco Strino, Boaz Nadler, and Yuval Kluger.
\newblock Ranking and combining multiple predictors without labeled data.
\newblock \emph{Proceedings of the National Academy of Sciences}, 111\penalty0 (4):\penalty0 1253--1258, 2014.

\bibitem[Cox et~al.(2015)Cox, Little, and O'Shea]{Cox}
D.~Cox, J.~Little, and D.~O'Shea.
\newblock \emph{Ideals, Varieties, and Algorithms: An Introduction to Computational Algebraic Geometry and Commutative Algebra}.
\newblock Springer-Verlag, 4th edition, 2015.

\bibitem[Corrada-Emmanuel(2023)]{corradaemmanuel2023streaming}
Andrés Corrada-Emmanuel.
\newblock Streaming algorithms for evaluating noisy judges on unlabeled data -- binary classification, 2023.

\end{thebibliography}

\end{document}